\documentclass{article}
\usepackage{ijcai25}

\usepackage{times}
\usepackage{soul}
\usepackage{url}
\usepackage[hidelinks]{hyperref}
\usepackage[utf8]{inputenc}
\usepackage[small]{caption}
\usepackage{graphicx}
\usepackage{amsmath}
\usepackage{amsthm}
\usepackage{booktabs}
\usepackage{algorithm}
\usepackage{algorithmic}
\usepackage[switch]{lineno}
\usepackage{amssymb}
\usepackage{pifont}
\usepackage{multirow}
\usepackage{adjustbox}
\usepackage{colortbl}
\usepackage{makecell}
\usepackage[table,xcdraw]{xcolor}

\newtheorem{theorem}{Theorem}
\newtheorem{define}{Definition}
\newtheorem{assume}{Assumption}

\title{Exploiting Label Skewness for Spiking Neural Networks in Federated Learning}

\author{
Di Yu$^1$\and
Xin Du$^1$\thanks{Corresponding Authors: Xin Du and Shuiguang Deng.}\and
Linshan Jiang$^2$\and
Huijing Zhang$^1$\And
Shuiguang Deng$^1$\\
\affiliations
$^1$Zhejiang University\\
$^2$National University of Singapore\\
\emails
\{yudi2023,xindu\}@zju.edu.cn, linshan@nus.edu.sg, \{huijingzhang, dengsg\}@zju.edu.cn
}

\begin{document}

\maketitle

\begin{abstract}
The energy efficiency of deep spiking neural networks (SNNs) aligns with the constraints of resource-limited edge devices, positioning SNNs as a promising foundation for intelligent applications leveraging the extensive data collected by these devices.
To address data privacy concerns when deploying SNNs on edge devices, federated learning (FL) facilitates collaborative model training by leveraging data distributed across edge devices without transmitting local data to a central server. 
However, existing FL approaches struggle with label-skewed data across devices, which leads to drift in local SNN models and degrades the performance of the global SNN model. In this paper, we propose a novel framework called FedLEC, which incorporates intra-client label weight calibration to balance the learning intensity across local labels and inter-client knowledge distillation to mitigate local SNN model bias caused by label absence.
Extensive experiments with \emph{three} different structured SNNs across \emph{five} datasets (i.e., \emph{three} non-neuromorphic and \emph{two} neuromorphic datasets) demonstrate the efficiency of FedLEC. Compared to \emph{eight} state-of-the-art FL algorithms, FedLEC achieves an average accuracy improvement of approximately 11.59\% for the global SNN model under various label skew distribution settings.
\end{abstract}

\section{Introduction}

Recent research has demonstrated that deep spiking neural networks (SNNs) are well-suited for deployment on resource-constrained edge devices \cite{di2024ec}, thanks to their energy efficiency \cite{maass1997networks} and compatibility with the computational limitations of these devices. 
These edge devices can also collect vast data, including photos, videos, and audio, which presents numerous opportunities with deep SNNs for influential research and practical applications \cite{lim2020federated}.
However, the traditional approach of centralizing this data has become progressively impractical, as it often involves sensitive information unsuitable for sharing, and the communication costs of transmitting large volumes of data are prohibitively high.

Federated Learning (FL) has emerged as a viable solution, enabling the development of deep SNN models while keeping data localized on edge devices \cite{venkatesha2021federated}. 
In FL, edge devices, commonly called \textit{clients}, train \textit{local models} on their respective local datasets. These locally trained models are subsequently transmitted to the central FL \textit{server}, aggregating them into a \textit{global model} to facilitate real-world applications. However, the characteristics of the data distribution across edge devices can present significant challenges to federated SNN learning, including:

\paragraph{Label Skewness.}
This type of data heterogeneity is common in practice as edge devices may only possess a subset of the global categories in their acquired datasets according to their deployed environments. For example, the distribution of animal species varies across different geographical regions, such as pandas being native to China and koalas exclusive to Australia.
When the corresponding label distributions across clients are extremely \textbf{imbalanced}, local models struggle to generalize to the global distribution, inducing a sub-optimal global model \cite{xia2024flea}.

\paragraph{Bias Aggravation.}
Although back-propagation with surrogate functions \cite{wu2019direct} allows local SNN models to be efficiently trained by edge GPUs, it also introduces layer-wise accumulated gradient drifts \cite{deng2023surrogate}. 
When SNNs are trained on highly skewed data, the absence of relevant knowledge about \textbf{missing labels} aggravates training drift, leading to severe gradient bias and increased local over-fitting. Hence, aggregating biased gradients from clients causes the global SNN model to converge poorly, ultimately degrading its performance \cite{wang2023adaptive}.

To mitigate these issues, we propose a novel FL framework named FedLEC\footnote{https://github.com/AmazingDD/FedLEC}.
\emph{\textbf{First}}, FedLEC introduces a calibrated objective function based on the corresponding probability of local label occurrences. By directing the training process to highlight the margins of majority and minority labels to achieve optimal thresholds, FedLEC promotes larger margin attainment for underrepresented minority labels, thereby enhancing their representation within the model and balancing the intra-client inconsistency. 
\emph{\textbf{Second}}, we employ a label distillation method to convey global distribution information regarding missing labels across clients. 
This approach ensures the local outputs align closely with the global model's predictions for missing labels, effectively mitigating the aggravated bias and over-fitting issues arising from label absence.
Besides, FedLEC focuses on alleviating the impact of locally skewed data so that it can be integrated with other methods \cite{karimireddy2020scaffold,li2020federated} that address inter-client objective inconsistency to improve the performance of the server model further.

To validate the effectiveness of our FedLEC, we conduct extensive experiments under various label skew settings,  substantiating that FedLEC significantly improves the accuracy of federated SNN learning compared to other state-of-the-art (SOTA) FL algorithms under label skews, particularly under extreme label skew conditions. The primary contributions of this paper are summarized as follows:
\begin{itemize}
    \item We first analyze the impact of label skewness in federated SNN learning and demonstrate that applying FL methods designed for artificial neural networks (ANNs) under extreme label skew conditions is suboptimal. This analysis lays the foundation for developing specialized solutions tailored to SNNs.
    \item We propose FedLEC, a federated learning framework designed to address the challenges posed by extreme label skew. FedLEC enhances the generalization ability of local SNN models by balancing intra-client label learning and mitigating bias from missing labels through inter-client knowledge alignment. This improves the global model's overall performance.
    \item Results from extensive experiments with \textbf{three} differently structured SNNs and \textbf{eight} FL baselines across \textbf{five} datasets demonstrate the efficacy of FedLEC. Under different extreme label skews, FedLEC can outperform the other baselines on average accuracy by about 11.59\% across all empirical trials in this study.
\end{itemize}

\section{Related Work}

Researchers have investigated how to leverage the energy efficiency of SNNs in FL systems \cite{venkatesha2021federated}. 
Previous studies mainly focused on federated SNN learning across various application scenarios.
For instance, FedSNN-NRFE \cite{xie2022efficient} exploited privacy-preserving FL, training SNN models for traffic sign recognition on the Internet of Vehicles. 
A distributed FL system with SNNs \cite{zhang2024federated} was built to process radar data collaboratively. 
SURFS \cite{aouedi2024surfs} proposed a robust and sustainable instruction detection system with federated SNN learning. 
Some studies also explore the compatibility of different FL architectures with SNNs. 
Hierarchical FL \cite{aouedi2023hfedsnn,aouedi2024surfs} was utilized with SNNs to lower the communication latency and enhance model robustness.

However, these studies overemphasize the application implementation by combining FL with SNNs and neglect the inherently heterogeneous data risks in real-world systems. 
Although several studies \cite{venkatesha2021federated,tumpa2023federated} have preliminarily explored the impact of non-IID data on federated SNN learning, the influence of one common non-IID data heterogeneous challenge called label skew is often neglected by most researchers.
Meanwhile, few studies have investigated whether solutions for label skews in FL applied to ANNs like FedLC \cite{zhang2022federated} and FedConcat \cite{diao2024exploiting} remain the same effectiveness for SNNs.
Motivated by current research limitations, this paper delves into tackling the significant drop in accuracy performance observed when implementing federated SNN learning under label skew.

\section{Preliminary}

\subsection{Federated Learning}

An FL system comprises $M$ local client nodes and a central global server. 
The server initiates the federated training process by broadcasting the initial model parameters $\theta$ to selected client nodes. 
Subsequently, the locally trained model at each client $m$ is updated using the client's private data shard $D^m$.
The model update from each client comprises the accumulated gradients throughout local training. These updates are periodically communicated to the server for aggregation. 
In one communication round, the updated parameters $\theta^m$ from client $m$ are transmitted to the server for global model updating. The server will then aggregate these perceived parameters in a specific manner to update the global model. 
A standard aggregation method using the weighted average is FedAvg \cite{mcmahan2017communication}.

\subsection{Spiking Neural Network}
 
\textit{Leaky Integrate-and-Fire} (LIF) neuron, widely used in diverse domains \cite{yao2024spikedriven,su2023deep}, is one of the popular spiking neurons utilized to build spiking neural networks, whose computational paradigm is:
\begin{align}
     V[t^-] & = V[t-1] + \beta(I[t] + V[t-1]) \\ \label{equ:heaviside}
     S[t] & = H(V[t^-]- \overline{V}) \\ 
     V[t] &= S[t]V_r + (1-S[t])V[t^-]
\end{align}
where $\beta, \overline{V}$, and $V_r$ represent the membrane time constant, the firing threshold, and the reset membrane potential, respectively. At time step $t$, $I[t]$ is the spatial input, $V[t^-]$ and $V[t]$ are the membrane potential after neuronal dynamics integration and after the trigger of firing separately. $H(\cdot)$ is the Heaviside step function, generating a binary spike when $V[t^-] \geq \overline{V}$. Subsequently, $V[t]$ will reset to $V_r$ or remain unchanged otherwise. To tackle the non-differentiable issue of Equation~(\ref{equ:heaviside}), studies \cite{wu2019direct,deng2023surrogate} introduce back-propagation through time (BPTT) with surrogate functions to train SNNs efficiently on GPUs. 

\section{Methodology}

\subsection{Problem Statement} 

\begin{figure}[!t]
\centering
\includegraphics[width=\columnwidth]{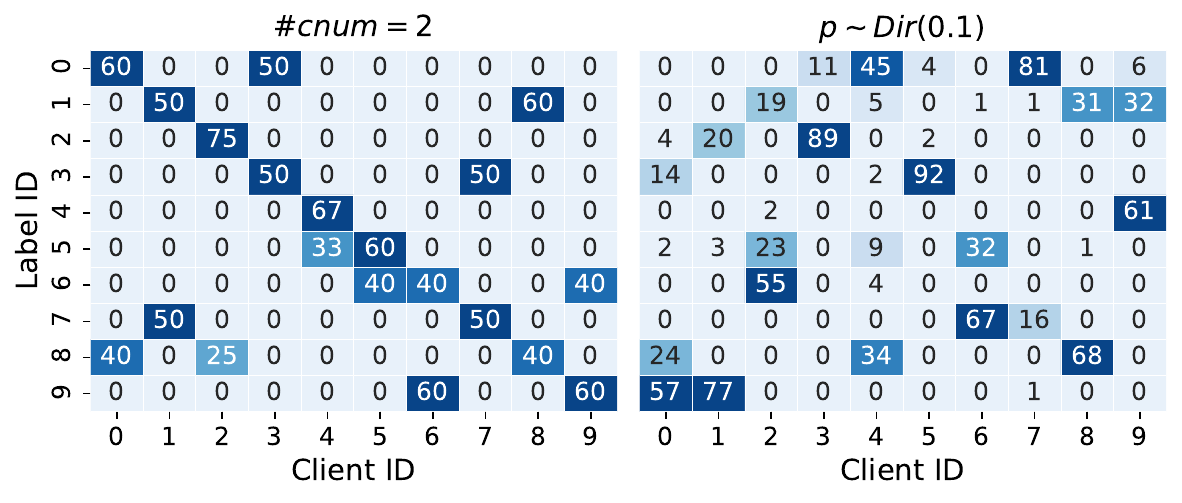} 
\vspace{-0.2in}
\caption{Different types of label skews. Cell number represents the percentage value of the samples of a specific class allocated.}
\vspace{-0.1in}
\label{fig:labelskew}
\end{figure}

There are two different label skew settings in FL \cite{li2022federated}: \textit{quantity-based} and \textit{distribution-based}, as depicted in Figure~\ref{fig:labelskew}.
For \textit{quantity-based} label skew, each client can only hold samples of fixed $c$ different labels. Samples with a specific label will be randomly divided into equal data shards and distributed to clients who own this label so that no overlap exists among the data shards of different clients (for simplicity, we use $\#cnum=c$ to denote this skew). 
The other type of label imbalance is \textit{distribution-based label skew} \cite{yurochkin2019bayesian}, which allocates portions of the samples for each label according to Dirichlet distribution \cite{huang2005maximum}. In particular, this skew allocates $p_{c,m}\%$ samples of class $c$ to client $m$, where $p_c \sim Dir(\alpha)$ and $\alpha\ (>0)$ is the concentration parameter to control the imbalance level. In this study, we utilize $p \sim Dir(\alpha)$ to represent this type of label skew. 
\begin{define}[Majority, Minority and Missing Labels]
    For a label-skewed local data shard $D$, its label set \(\mathcal{C}\) can be divided into three categories: the majority label set \(\mathcal{J}\), the minority label set \(\mathcal{K}\), and the missing label set \(\mathcal{M}\). We have \(\mathcal{C} = \mathcal{J} \cup \mathcal{K} \cup \mathcal{M}\), \(\mathcal{J} \cap \mathcal{K} = \emptyset\), \(\mathcal{J} \cap \mathcal{M} = \emptyset\), \(\mathcal{M} \cap \mathcal{K} = \emptyset\), where $\emptyset$ is an empty set. The number of samples $(\mathbf{x}, y)\sim D$ with label set $\mathcal{C}$ satisfies \(|D_\mathcal{J}| \gg |D_\mathcal{K}| > |D_\mathcal{M}| = 0\).
\end{define}
In FL, a total of $M$ clients aim to minimize the below objective with model parameter $\theta$:
\begin{align}\label{eq:global-loss}
    \underset{\theta\in \mathbb{R}^d}{\mathrm{min}}\ \mathcal{L}(\mathbf{\theta}) = \sum_{m=1}^M \mathcal{P}_m \cdot \mathcal{L}_m(\theta)
\end{align}
where $\mathcal{L}_m$ is the objective function for local model in the $m$-th client, $\mathcal{P}_m$ is the relative local sample size, and $\sum_{m=1}^M\mathcal{P}_m=1$. The local data distribution $P_m$ varies across clients for label skew settings, resulting in significantly different models after local training. 

\begin{figure}[!t]
\centering
\includegraphics[width=\columnwidth]{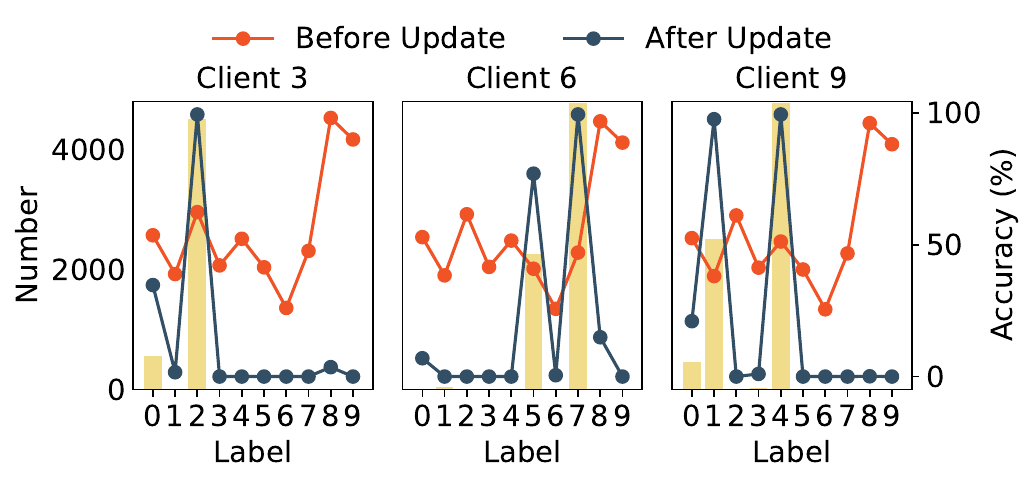} 
\vspace{-0.2in}
\caption{The impact of label skewness on local updating for models in three selected clients. The histogram indicates the number of samples for each label in the local data shard.}
\vspace{-0.1in}
\label{fig:twins}
\end{figure}

Besides, we empirically validate the impact of label skew in federated SNN learning on three selected clients with \textit{cifar10} dataset after sufficient communication rounds. 
After verifying that the current local models are consistent with the global SNN model, we evaluate the accuracy for each label both before and after the local update.
As shown in Figure~\ref{fig:twins}, after local training is conducted separately, the test accuracy for majority labels typically improves compared to the initial state. However, the test accuracy for minority and missing labels significantly decreases, often approaching zero. This observation highlights that label skewness can result in a biased SNN model
. 

\subsection{Framework Overview}

\begin{figure}[!t]
\centering
\includegraphics[width=\columnwidth]{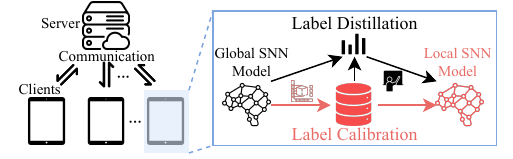} 
\vspace{-0.2in}
\caption{The framework of FedLEC.}
\vspace{-0.1in}
\label{fig:fedlec}
\end{figure}

We propose a novel \textbf{fed}erated SNN learning framework with \textbf{l}abel sk\textbf{e}w \textbf{c}alibration (FedLEC).
As shown in Figure~\ref{fig:fedlec}, we conduct two strategies to alleviate the intra-client label inconsistency. 
\emph{\textbf{First}}, we calibrate the logits\footnote{Logits refer to the output of the model's final classification layer} 
 of each label before applying the softmax cross-entropy by adjusting them based on the occurrence probability of each label, thereby mitigating the impact of intra-client data imbalance from minority labels during the local update.
\emph{\textbf{Second}}, the global model from the previous communication round, serving as a teacher model, transfers knowledge about the inter-client data distribution to each local model, thereby mitigating over-fitting issues related to missing labels.
Then, each local objective function $\mathcal{L}_m$ in Equation~(\ref{eq:global-loss}) for one time step is:
\begin{align} \label{eq:totalloss}
    \underset{\theta \in \mathbb{R}^d}{\mathrm{min}}\ \mathcal{L}_m(\theta)= (1-\lambda) \cdot \mathcal{L}_c(\theta) + \lambda\mathcal{L}_d(\theta)
\end{align}
where $\mathcal{L}_c$ and $\mathcal{L}_d$ represent the objective functions derived from the two strategies, respectively. 
By adjusting the hyper-parameter $\lambda$,  the model balances its focus between addressing intra-client and inter-client bias during local learning.

\paragraph{Label Calibration.} 
Denote the data distribution at $m$-th client as $P(\mathbf{x}, y)=P(\mathbf{x}|y)P(y)$. For a sample $\mathbf{x}$, the predicted label is $\hat{y}=\mathrm{arg\ max}_yf(\mathbf{x};\theta)$, where $f(\mathbf{x};\theta)$ is the corresponding logits from the local model. 
In the one communication round of the FL process, the server broadcasts the global model parameters to the selected local client $m$.
The goal of the local training task is to minimize the classification error from a statistical perspective:
\begin{align}\label{eq:loss}
    \mathrm{min}\ P(y\neq \hat{y}|\mathbf{x}) 
\end{align}
According to Bayes rule, Equation~(\ref{eq:loss}) is proportional to $P(\mathbf{x}|y)P(y)$. When the label distribution is balanced, i.e., $P(y)$ is equivalent for all label $y$, then the soft-max cross-entropy can be regarded as the surrogate loss function for Equation~(\ref{eq:loss}) since the probability $P(\mathbf{x}|y) \propto e^{f(\mathbf{x;\theta})}$.
However, when the local data suffer from label distribution skewness, $P(y)$ is no longer identical, i.e., minority labels have much lower probabilities of occurrence than majority labels, inspiring us to calibrate the error in Equation~(\ref{eq:loss}) via averaging the per-label error rate \cite{menon2021longtail} like: $\frac{1}{|\mathcal{C}|}\sum_{y\in \mathcal{C}}P(\mathbf{x}|y\neq\hat{y})$. 
In this manner, varying $P(y)$ due to label skewness will no longer affect the results.
To minimize the `balanced' calibrated error, we, in turn, need to find:
\begin{align}
\begin{aligned}\label{eq:lc}
    & \mathrm{arg\ max}_{y\in\mathcal{C}}P(\mathbf{x}|y=\hat{y})= \mathrm{arg\ max}_{y\in\mathcal{C}} P(y|x)/P(y) \\ 
    & = \mathrm{arg\ max}_{y\in\mathcal{C}} (f_y(\mathbf{x};\theta) - \mathrm{log}(\gamma_y)) 
\end{aligned}
\end{align}
where $\gamma_y$ is the estimation of the prior $P(y)$ for a specific label in the local label set $\mathcal{C}$. The logits of minority labels in $\mathcal{C}$ are reduced by a relatively larger value, whereas those of majority labels are adjusted oppositely, thereby balancing the classification loss. Inspired by Equation~(\ref{eq:lc}), we can then alleviate the impact of intra-client label inconsistency by training the local SNN model with a calibrated cross-entropy loss:
\begin{align}
    \mathcal{L}_c= -\mathbb{E}_{(\mathbf{x},y)\sim D}\mathrm{log} \left( \frac{e^{f_y(\mathbf{x};\theta)} - \gamma_y}{\sum_{c\in \mathcal{C}} (e^{f_c(\mathbf{x};\theta)} - \gamma_c)}  \right)
\end{align}
Notably, $\mathcal{L}_c$ prioritize mitigating the biases between the majority and minority labels. However, it fails to handle missing label issues since no related class prior is available. Hence, it is necessary to design distinct objective functions to ensure that the global model retains its original predictive capability for missing labels, preventing the loss of such capability due to over-fitting caused by the absence of relevant data.

\paragraph{Label Distillation.} Figure~\ref{fig:labelskew} has demonstrated that extreme quantity-based and distribution-based label skews often induce numerous labels missing in local data shard, i.e., $|\mathcal{M}|\gg |\mathcal{J}| + |\mathcal{K}|$. After implementing local training, each client's local model will be over-fitted according to the local data distribution. 
The core idea to alleviate this issue is to \emph{preserve the global view of the data distribution for locally missing labels}. 
Specifically, when the server aggregates gradients from the local models during each communication round, the global could learn and retain distribution information about all the labels. 
Therefore, we employ the predictions of the global model on locally available data to perform knowledge distillation (KD) for local missing labels in $\mathcal{M}$. 
Besides, KD can effectively mitigate the over-fitting issue of SNNs caused by accumulated gradient errors \cite{zuo2024self}. Then, FedLEC defines the distillation loss $\mathcal{L}_d$ as the KL-divergence loss between the soft-max prediction vector of the global model $\Tilde{q}_g$ and that of the local model $\Tilde{q}_l$ as follows:
\begin{align}
    \mathcal{L}_d= \mathbb{E}_{(\mathbf{x},y)\sim D}\sum_{c\in \mathcal{M}} \Tilde{q}_g[c] \mathrm{log}\frac{\Tilde{q}_g[c]}{\Tilde{q}_l[c]}
\end{align}
which guarantees the local model effectively aligns with the global model's predictive capabilities for missing labels.

\paragraph{Aggregation Through Time.} 
According to Equation~(\ref{eq:global-loss}), the aggregate process of SNN model parameters in FedLEC at one communication round can be expressed as: 
\begin{align}\label{eq:t-fedavg}
    \theta & = \underbrace{\sum_{m=1}^M \mathcal{P}_m \left( \frac{1}{T}\sum_{t=1}^T\theta^t \right)}_{\text{FedLEC}} = \frac{1}{T}\sum_{t=1}^T \underbrace{(\sum_{m=1}^M \mathcal{P}_m \theta^t) }_{\text{FedAvg}}
\end{align}
where $T$ is the total time steps the SNN model training process requires. 
Intuitively, the global aggregation process of FedLEC scales with time steps compared to conventional FedAvg. We term this parameter aggregation process in federated SNN learning as Aggregation Through Time (ATT), whose workflow is elaborated in Algorithm~\ref{alg:fedavg}. 

\begin{algorithm}[!t]
\caption{Aggregation Through Time in FedLEC}
\label{alg:fedavg}
\raggedright \textbf{Input}: Selected Client Set $\mathcal{S}$, Total Timesteps $T$ \\
\raggedright \textbf{Output}: Updated SNN Parameters $\theta$. \\
\begin{algorithmic}[1] 
\STATE \textbf{Server executes}: 
\FORALL{$t=1,...,T$}
    \STATE Receive parameters $\theta^{m,t}$ from each client $m$.
    \STATE $\theta^t \leftarrow  \sum_{m\in \mathcal{S}} \frac{|D^m|}{\sum_{m\in \mathcal{S}}|D^m|} \theta^{m,t}$ 
\ENDFOR
\STATE $\theta \leftarrow \frac{1}{T}\sum_{t=1}^T \theta^t$.
\STATE \textbf{return} $\theta$
\end{algorithmic}
\end{algorithm}

\subsection{Convergence Analysis}

Inspired by \cite{li2019convergence}, we make the following assumptions on the objectives $\mathcal{L}_1, ..., \mathcal{L}_M$. Assumptions 1 and 2 are standard; a typical example is the soft-max classifier. Since the KL-divergence loss $\mathcal{L}_d$ and the calibrated cross-entropy $\mathcal{L}_c$ are typical variants of the soft-max classifier, the local objective $\mathcal{L}_m$ of FedLEC also satisfies these assumptions.
\begin{assume}[Each $\mathcal{L}_m$ is $L$-smooth]
    For all $\mathbf{v}$ and $\mathbf{w}$, $\mathcal{L}_m(\mathbf{v}) \leq \mathcal{L}_m(\mathbf{w}) + (\mathbf{v}-\mathbf{w})^T\nabla \mathcal{L}_m(\mathbf{w})+\frac{L}{2}||\mathbf{v} - \mathbf{w}||_2^2$
\end{assume}
\begin{assume}[Each $\mathcal{L}_m$ is $\mu$-strongly convex]
    For all $\mathbf{v}$ and $\mathbf{w}$, $\mathcal{L}_m(\mathbf{v}) \geq \mathcal{L}_m(\mathbf{w}) + (\mathbf{v}-\mathbf{w})^T\nabla \mathcal{L}_m(\mathbf{w}) + \frac{\mu}{2}||\mathbf{v} - \mathbf{w}||_2^2$
\end{assume}
Let the number of local iterations performed in one device between two communications be $E$, and the total number of every device's SGDs be $R$. 
Given that Equation~(\ref{eq:t-fedavg}) ensures $R$ is evenly divisible by the number of time steps $T$, we simplify the subsequent analysis by omitting $T$, as its effect is already encapsulated within $R$.
Note that with surrogate functions \cite{wu2019direct}, SNNs can implement back-propagation with gradient descent like ANNs. Hence, for the $r$-th SGD of the local SNN model in client $m$, we have:
\begin{assume}[Variance boundary]
    Let $\xi_r^m$ be randomly sampled from the $m$-th client's local data. Each device's variance of stochastic gradients is bounded: $\mathbb{E}|| \nabla \mathcal{L}_m(\theta_r^m, \xi_r^m) - \nabla\mathcal{L}_m(\theta_r^m) ||^2 \leq \sigma^2_m$, for $m=1, ..., M$.
\end{assume}
\begin{assume}
    The expected squared norm of stochastic gradients is uniformly bounded, i.e., $\mathbb{E}|| \nabla\mathcal{L}_m(\theta_r^m, \xi_r^m) ||^2 \leq G^2$ for all $m=1,..,M$ and $r=1,...,R-1$.
\end{assume}
Let $\mathcal{L}^*$ and $\mathcal{L}_m^*$ be the minimum values of $\mathcal{L}$ and $\mathcal{L}_m$, respectively. We utilize the term $\Gamma=\mathcal{L}^*-\sum_{m=1}^M\mathcal{P}_m\mathcal{L}^*_m$ to quantify the degree of data heterogeneity (label skewness). If the data is IID, $\Gamma$ approaches zero as the number of samples increases. If the data is non-IID, then $\Gamma$ is nonzero, and its magnitude reflects the skewness of the data distribution. 

We first analyze the case that all the clients participate in the parameter aggregation step. From Equation~(\ref{eq:t-fedavg}), we acknowledge that the aggregation step in FedLEC for each time step $t$ is equivalent to that of FedAvg, denoted as: 
\begin{align}
    \theta_{r+i+1}^m \leftarrow \theta^m_{r+i} - \eta_{r+i}\nabla \mathcal{L}_m(\theta^m_{r+i}, \xi^m_{r+i})
\end{align}
where $\eta_{t+i}$ is the learning rate (a.k.a. step size) and $i=0,1,...,R-1$. After $R$ iterations, FedLEC terminates and returns the latest aggregated global model parameters $\theta_{R}$ as the FedLEC solution, then we have:
\begin{theorem}
    Let Assumptions 1 to 4 hold and $L$, $\mu$, $\sigma_m$, $G$ be defined therein. Choose $\kappa=\frac{L}{\mu}$, $\gamma=\mathrm{max}\{8\kappa, E \}$ and the learning rate $\eta_t=\frac{2}{\mu(\gamma + r)}$. Then FedLEC with full device participation satisfies:
    \begin{equation}
        \resizebox{.88\linewidth}{!}{$
            \displaystyle
            \mathbb{E}[\mathcal{L}(\theta_R)]-\mathcal{L}^*\leq \frac{\kappa}{\gamma + R - 1}\left( \frac{2B}{\mu} + \frac{\mu\gamma}{2} \mathbb{E}||\theta_0 - \theta^*||^2 \right)
        $}
    \end{equation}%
    where
    \begin{align}\label{eq:B}
        B = \sum_{m=1}^M\mathcal{P}_m^2\sigma_m^2+6L\Gamma+8(E-1)^2G^2
    \end{align}
\end{theorem}
If a random set $\mathcal{S}$ of clients is selected randomly without replacement to participate in aggregating, $|\mathcal{S}|=K$, the aggregation step of FedLEC performs $\theta_r\leftarrow\frac{N}{K}\sum_{m\in \mathcal{S}}\mathcal{P}_m\theta_r^m$.
By converting the FL objective in Equation~(\ref{eq:global-loss}) as: $\mathcal{L}(\theta) = \frac{1}{M}\sum^M_{m=1}\Tilde{\mathcal{L}}_m(\theta)$, where $\Tilde{\mathcal{L}}_m(\theta)=\mathcal{P}_mM\mathcal{L}_m(\theta)$, we have:
\begin{theorem}
    Let Assumptions 1 to 5 hold and define $\Tilde{C}=\frac{M-K}{M-1}\frac{4}{K}E^2\Tilde{G}^2$, then
    \begin{equation}\label{eq:partial}
        \resizebox{.88\linewidth}{!}{$
            \displaystyle
            \mathbb{E}[\mathcal{L}(\theta_R)]-\mathcal{L}^*\leq \frac{\kappa}{\gamma + R - 1}\left( \frac{2(\Tilde{B}+\Tilde{C})}{\Tilde{\mu}} + \frac{\Tilde{\mu}\gamma}{2} \mathbb{E}||\theta_0 - \theta^*||^2 \right)
        $}
    \end{equation}%
    where $\Tilde{B}$ is similar to Equation~(\ref{eq:B}), but $L, \mu, \sigma, G$ are replaced by $\Tilde{L}\triangleq \nu L, \Tilde{\mu}\triangleq \varsigma\mu, \Tilde{\sigma}\triangleq \sqrt{\nu} \sigma$, and $\Tilde{G}\triangleq \sqrt{\nu}G$, respectively. Here, $\nu=M\cdot \mathrm{max}_m \mathcal{P}_m$ and $\varsigma = M\cdot \mathrm{min}_m \mathcal{P}_m$
\end{theorem}
Since $||\theta_0 - \theta^*||^2 \leq \frac{4}{\mu^2}G^2$ for $\mu$-stronly convex $\mathcal{L}$, the dominating term in Equation~(\ref{eq:partial}) is:
\begin{align}\label{eq:bigo}
    \mathcal{O}\left( \frac{\sum_{m=1}^N \mathcal{P}_m^2\sigma^2_m + L\Gamma + (1 + \frac{1}{K})E^2G^2 + \gamma G^2}{\mu R} \right)
\end{align}
Given the pre-defined settings in Assumptions 1 to 4, we conclude that FedLEC will converge with $\mathcal{O}(\frac{1}{R})$ rate in non-IID FL settings like label skewness according to Equation~(\ref{eq:bigo}).

\subsection{Privacy Protection}
FedLEC inherits the privacy-preserving merit of the FL framework, which allows each client to preserve private data locally and significantly reduce the risk of privacy leakage.
To further handle the potential privacy violation when uploading model parameters to the server, we propose integrating the local differential privacy strategy \cite{el2022differential} into FedLEC. Specifically, we incorporate a zero-mean Laplacian noise to the SNN model parameters before it is uploaded to the server:
\begin{align}
    \theta^m = \theta^m + \mathrm{Laplacian}(0, \delta)
\end{align}
where $\delta$ is the noise intensity. Hence, one cannot easily obtain the updated data by monitoring the parameter change, and the privacy protection ability is better as $\delta$ increases.

\section{Experiments}

\subsection{Experimental Settings}
In this study, we use \textbf{four} FL algorithms designed for tackling conventional data heterogeneity, i.e., FedAvg, FedProx \cite{li2020federated}, FedNova \cite{wang2020tackling}, and Scaffold \cite{karimireddy2020scaffold}), and \textbf{four} extra FL algorithms proposed to tackle the label skewness, i.e., FedLC \cite{zhang2022federated}, FedRS \cite{li2021fedrs}, FLea \cite{xia2024flea}, and FedConcat \cite{diao2024exploiting}, as compared baselines to demonstrate the effectiveness of FedLEC. 
All FL algorithms adopt \textit{S-VGG9} \cite{venkatesha2021federated} as the default backbone SNN model.

We uniformly execute 50 communication rounds, selecting 20\% parts from 10 clients to train their models for 10 local epochs per round with a batch size of 64 as default. We use the Adam optimizer for all trials with a learning rate 0.001. 
All experimental trials are implemented on NVIDIA GeForce RTX 4090 GPUs.

\begin{table}[!t]
\centering
\begin{adjustbox}{width=\columnwidth}
\begin{tabular}{llrrrrr}
\toprule
Dataset & Partition & FedAvg & FedProx & Scaffold & FedNova & FedLEC \\ 
\midrule
\multirow{5}{*}{\textit{cifar10}} 
 & $IID$ & 83.91 & 82.53 & \textbf{85.18} & 79.89 & \cellcolor{gray!20}$-$ \\
  & $p\sim Dir(0.05)$ & \underline{35.80} & 32.16 & 29.42 & 27.37 & \cellcolor{gray!20}(\textcolor{red}{$\uparrow 11.07$}) \textbf{46.87} \\ 
  & $p\sim Dir(0.1)$ & 53.45 &  51.76 & \underline{54.08}  & 47.69 & \cellcolor{gray!20} (\textcolor{red}{$\uparrow 13.56$}) \textbf{67.64} \\
 & $\#cnum=2$ & 29.60 & 24.64 & \underline{34.65} & 25.61 & \cellcolor{gray!20}(\textcolor{red}{$\uparrow\ \ 6.51$}) \textbf{41.16} \\
 & $\#cnum=4$ & 54.24 & 56.94 & \underline{64.28} & 53.82 & \cellcolor{gray!20}(\textcolor{red}{$\uparrow\ \ 4.96$}) \textbf{69.24} \\

\midrule
	
 \multirow{5}{*}{\textit{cifar100}} 			
 & $IID$ & 52.78 & 46.17 & \textbf{54.56} & 33.37 & \cellcolor{gray!20}$-$ \\
 & $p\sim Dir(0.05)$ & 24.11 & 25.01 & \underline{26.11} & 13.90 & \cellcolor{gray!20}(\textcolor{red}{$\uparrow 14.46$}) \textbf{40.57} \\
  & $p\sim Dir(0.1)$ & 29.29 & 29.45 & \underline{33.06} & 15.51 & \cellcolor{gray!20}(\textcolor{red}{$\uparrow\ \ 9.67$}) \textbf{42.73} \\
 & $\#cnum=20$ & 20.80 & 22.28 & \underline{23.01} & 13.98 & \cellcolor{gray!20}(\textcolor{red}{$\uparrow 14.24$}) \textbf{37.25} \\
 & $\#cnum=40$ & 32.85 & 34.41 & \underline{37.23} & 20.85 & \cellcolor{gray!20}(\textcolor{red}{$\uparrow\ \ 9.84$}) \textbf{47.07} \\
 
\midrule

\multirow{5}{*}{\textit{svhn}} 
 & $IID$ & 93.94 & 93.28  & \textbf{94.26} & 91.03 & \cellcolor{gray!20} $-$ \\  
 & $p\sim Dir(0.05)$ & 44.53 & \underline{46.80} & 40.24 & 33.87 & \cellcolor{gray!20}
(\textcolor{red}{$\uparrow 12.06$}) \textbf{58.86} \\ 
  & $p\sim Dir(0.1)$ & 66.04 & \underline{68.26} & 66.48 & 67.82 & \cellcolor{gray!20}(\textcolor{red}{$\uparrow 14.70$}) \textbf{82.96} \\
 & $\#cnum=2$ & \underline{39.42} & 31.01 & 32.00 & 37.33 & \cellcolor{gray!20}(\textcolor{red}{$\uparrow\ \ 9.87$}) \textbf{49.29} \\
 & $\#cnum=4$ & 65.61 & 69.87 & \underline{72.66} & 69.02 & \cellcolor{gray!20}(\textcolor{red}{$\uparrow\ \ 9.44$}) \textbf{82.10} \\
 
 \bottomrule
\end{tabular}
\end{adjustbox}
\caption{Test accuracy (\%) of different federated SNN learning approaches w.r.t label skews. The $IID$ results are set as a reference, indicating how extreme label skews affect final accuracy. Typically, The skewness of $\#cnum=20/40$ for 100-labels is equivalent to $\#cnum=2/4$ for 10-labels. \textbf{Bold} value is the best result across all FL algorithms, while \underline{underline} value is the second-best. We run \textit{three} trials and report the mean top-$1$ accuracy.}
\label{tab:fl-snn-acc}
\end{table}

\subsection{Performance Evaluation}

\paragraph{Overall Accuracy.}
Table~\ref{tab:fl-snn-acc} presents the accuracy performance of FedLEC compared to conventional FL algorithms tailored for data heterogeneity. 
The results demonstrate the clear superiority of FedLEC in handling extreme label skew conditions. 
Specifically, in \textit{cifar10} dataset, FedLEC reaches an average improvement of about 9.03\%, while in \textit{svhn}, the average accuracy increment rate is 11.52\%. 
When the task extends to more complex classification tasks such as \textit{cifar100}, FedLEC similarly exhibits notable performance improvements, highlighting its potential for adaptation to complex and large-scale tasks.
Additionally, FedLEC demonstrates potential advantages in handling distribution-based label skews, as these skews do not result in the absence of a significant portion of labels. Consequently, they do not induce severe information loss, unlike quantity-based skews.

\begin{table}[!t]
\centering
\begin{adjustbox}{width=\columnwidth}
\begin{tabular}{llrrrrr}
\toprule
Dataset & Partition & FedLC & FedRS & FLea & FedConcat & FedLEC \\ 
\midrule
\multirow{2}{*}{\textit{cifar10}} & $\#cnum=2$ & 23.96 & 24.28 & \underline{32.66} & 32.64 & \cellcolor{gray!20}(\textcolor{red}{$\uparrow\ \ 8.50$}) \textbf{41.16} \\
 & $p\sim Dir(0.05)$ & 32.94 & \underline{33.49} & 28.16 & 30.43 & \cellcolor{gray!20}(\textcolor{red}{$\uparrow 13.38$}) \textbf{46.87} \\  
 
 \midrule

\multirow{2}{*}{\textit{cifar100}} & $\#cnum=20$ & 19.88 & 8.01 & \underline{28.57} & 25.78 & \cellcolor{gray!20}(\textcolor{red}{$\uparrow\ 8.68$}) \textbf{37.25} \\
 & $p\sim Dir(0.05)$ & 21.96 & 12.82 & \underline{26.62} & 23.15 & \cellcolor{gray!20}(\textcolor{red}{$\uparrow\ 13.95$}) \textbf{40.57} \\  

 \midrule
 
 \multirow{2}{*}{\textit{svhn}} & $\#cnum=2$ & 31.69 & 31.71 & \underline{33.46} & 32.84 & \cellcolor{gray!20}(\textcolor{red}{$\uparrow 15.83$}) \textbf{49.29} \\  
 & $p\sim Dir(0.05)$ & 29.97 & 39.10 & 42.67 & \underline{45.63} &  \cellcolor{gray!20}(\textcolor{red}{$\uparrow 13.23$}) \textbf{58.86}\\ 
 
\bottomrule
\end{tabular}%
\end{adjustbox}
\caption{Test accuracy (\%) of FedLEC compared with the other baselines specially designed for addressing extreme label skews, denoted in the same way as Table~\ref{tab:fl-snn-acc}.}
\vspace{-0.2in}
\label{tab:acc-imp}
\end{table}

Table~\ref{tab:acc-imp} further provides a performance comparison between FedLEC and several recently proposed FL algorithms specifically designed to mitigate label skewness.
Observations reveal that these algorithms are ineffective in federated SNN learning settings, exhibiting an average accuracy that is 12.26\% lower than that of FedLEC. 

The superiority of FedLEC may stem from the \textbf{fundamental differences} in training paradigms between ANNs and SNNs. 
BPTT with surrogate functions induces the intrinsic gradient bias accumulated by layer-wise back-propagation \cite{deng2023surrogate}. Extreme label skewness aggravates this bias, leading to drifting errors in the features extracted by the SNN model.
In particular, algorithms like FLea and FedConcat heavily rely on features extracted by local models to enhance resistance to label skewness. However, this dependency may backfire, further amplifying the occurrence of errors.
The distillation loss $\mathcal{L}_d$ can effectively mitigate excessive gradient drift. Since the predictions of the previous global model retain partial information about the correct optimization direction, they can serve as a reference during the current phase of local model training, helping to prevent over-fitting.
Hence, it is reasonable that the performance of FedLEC surpasses the other baselines by an average of approximately 11.33\%.

\paragraph{Results on Other SNN Architectures.}

\begin{figure}[!t]
\centering
\includegraphics[width=\columnwidth]{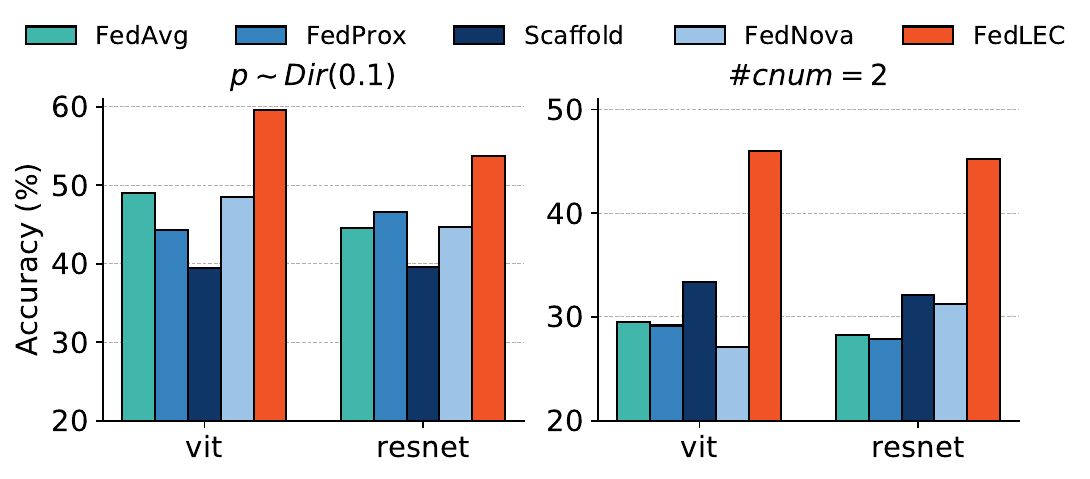} 
\vspace{-0.2in}
\caption{Accuracy performance of different models when varying various FL algorithms under different label skews.}
\vspace{-0.1in}
\label{fig:models-acc}
\end{figure}

To assess the generalization capability of FedLEC, we incorporate more advanced SNN models, such as \textit{MS-ResNet} \cite{hu2024advancing} and \textit{Meta-Spikeformer} \cite{yao2024spikedriven}, for training in an FL setting under various label skewness scenarios. 
For brevity, we simplify \textit{MS-ResNet} as \textit{resnet} and \textit{Meta-Spikeformer} as \textit{vit}. 
As depicted in Figure~\ref{fig:models-acc}, FedLEC achieves accuracy improvements of 10.13\% and 11.62\% for \textit{resnet} and \textit{vit}, respectively, demonstrating that the proposed local objective remains effective even for these complex SNN architectures.


\begin{figure}[!t]
\centering
\includegraphics[width=\columnwidth]{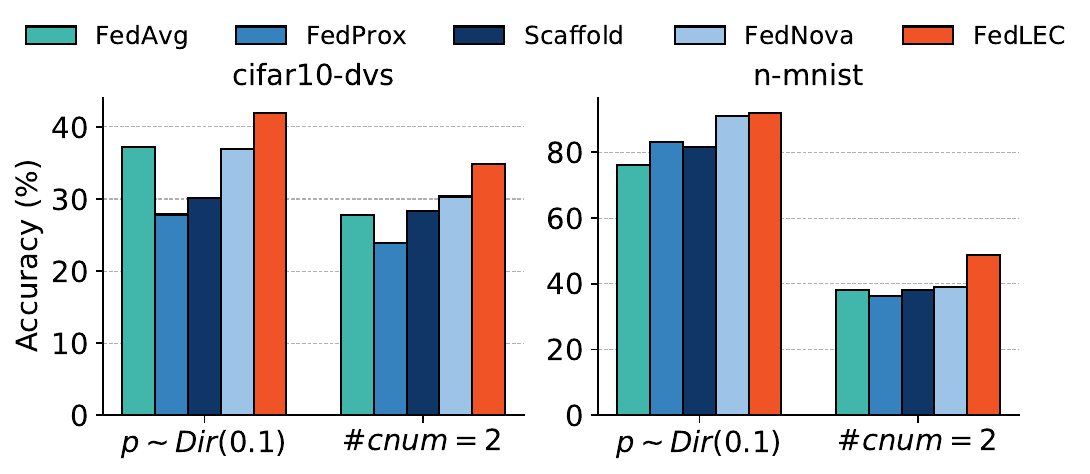} 
\vspace{-0.2in}
\caption{Accuracy performance of various federated SNN learning algorithms across different data scenarios under different skewness.}
\vspace{-0.1in}
\label{fig:tasks-acc}
\end{figure}

\paragraph{Results on Event-Based Datasets.} 
We also apply FedLEC to event-based image classification tasks, a representative application scenario for SNNs, to evaluate its generalized effectiveness across a broader range of scenarios.
Figure~\ref{fig:tasks-acc} shows that while FedLEC continues to exhibit superior performance, the performance gaps between FedLEC and other algorithms are notably smaller than those observed in static image classification tasks, with an average improvement of only approximately 4\%.
In particular, the classification task on the event-based dataset is more complicated since the features are sparse and chaotic. 
Hence, achieving a well-performing global model capable of capturing accurate patterns becomes challenging with highly skewed datasets, which limits the performance improvement of FedLEC.

\subsection{Efficiency Evaluation}

\begin{figure}[!t]
\centering
\includegraphics[width=\columnwidth]{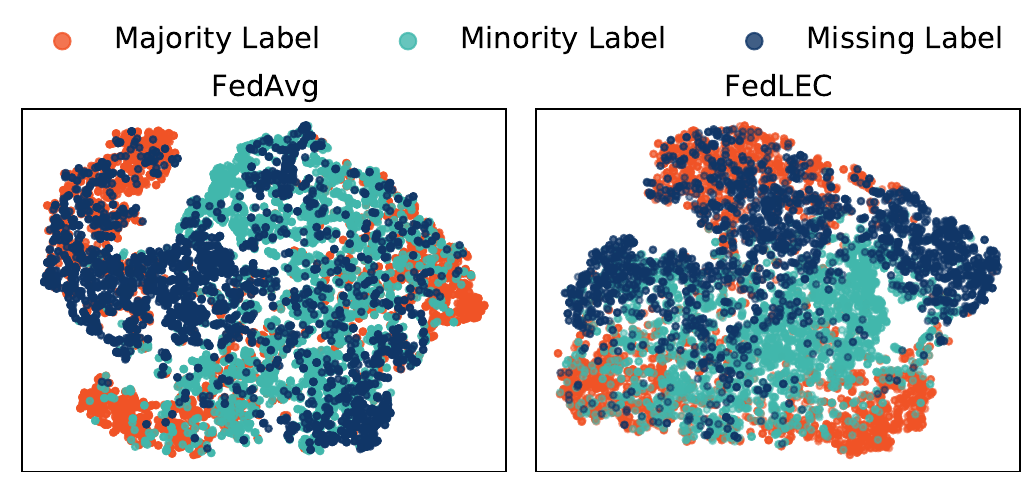} 
\vspace{-0.2in}
\caption{T-SNE visualizations on majority, minority, and missing labels. \textbf{Left:} For FedAvg, the samples from the minority and missing labels are mixed and indistinguishable. \textbf{Right:} For our method, the data from minority and missing labels can be distinguished well.}
\vspace{-0.1in}
\label{fig:tsne}
\end{figure}

\paragraph{T-SNE Visualization.}
We provide a visual analysis that examines the differences in the sample features extracted by the same models following local updates between FedLEC and FedAvg, aiming to demonstrate FedLEC's effectiveness.
We first randomly select a client with a local data shard where $|D_\mathcal{M}|=0$, $|D_\mathcal{K}|=249$, $|D_\mathcal{J}|=8437$. After the well-trained global models are locally trained in this client using FedAvg and FedLEC with one epoch, we feed them with the same test data and leverage the corresponding output features for visualization \cite{van2008visualizing}. 
As depicted in Figure~\ref{fig:tsne}, the test samples, especially those from minority (i.e., green samples) and missing (i.e., dark blue samples) labels, are mixed and difficult to distinguish. However, FedLEC can alleviate this issue and learn more discriminative features, reducing classification ambiguity when making inferences on samples from missing and minority labels.

\begin{table}[!t]
\centering
\begin{adjustbox}{width=\columnwidth}
\begin{tabular}{llrrrr}
\toprule
\multirow{2}{*}{$\mathcal{L}_c$} & \multirow{2}{*}{$\mathcal{L}_d$} & \multicolumn{2}{c}{\textit{cifar10}} & \multicolumn{2}{c}{\textit{svhn}} \\ \cmidrule(l){3-4} \cmidrule(l){5-6}
 &  & $p\sim Dir(0.15)$ & $\#cnum=3$ & $p\sim Dir(0.15)$ & $\#cnum=3$ \\ \midrule
\ding{55} & \ding{55} & 63.35 & 32.24 & 70.29 & 46.36  \\
\ding{51} & \ding{55} & (\textcolor{red}{$\uparrow 0.65$}) 64.00 & (\textcolor{red}{$\uparrow 3.06$}) 35.30 &  (\textcolor{red}{$\uparrow 2.89$}) 73.18 &  (\textcolor{red}{$\uparrow 0.42$}) 46.78 \\
\ding{55} & \ding{51} & (\textcolor{red}{$\uparrow 5.45$}) 68.80 & (\textcolor{red}{$\uparrow 24.6$}) 56.79 &  (\textcolor{red}{$\uparrow 13.7$}) 84.02 &  (\textcolor{red}{$\uparrow 26.7$}) 73.08 \\ \midrule
\rowcolor{gray!20}
\ding{51} & \ding{51} & (\textcolor{red}{$\uparrow 6.04$}) \textbf{69.39} &  (\textcolor{red}{$\uparrow 25.1$}) \textbf{57.30} &  (\textcolor{red}{$\uparrow 14.0$}) \textbf{84.24} &  (\textcolor{red}{$\uparrow 30.1$}) \textbf{76.49} \\ \bottomrule
\end{tabular}%
\end{adjustbox}
\caption{Impact of each loss component on Accuracy (\%).}
\label{tab:ablation}
\end{table}
\paragraph{Ablation Study.}
The results in Table~\ref{tab:ablation} quantify the contribution of each local loss term in FedLEC. From these results, we conclude that both terms significantly contribute to the final accuracy, with $\mathcal{L}_d$ being the dominant factor, leading to an average improvement of 17.61\%. 
In addition, the combination of both loss terms further enhances the federated SNN learning performance under extreme label skew, yielding an additional 1.20\% improvement. This improvement is attributed to $\mathcal{L}_d$ mitigating gradient biases and conveying the global data distribution, while $\mathcal{L}_c$ helps alleviate the distribution imbalance.

\subsection{Hyper-Parameter Analysis}
\begin{figure}[!t]
\centering
\includegraphics[width=\columnwidth]{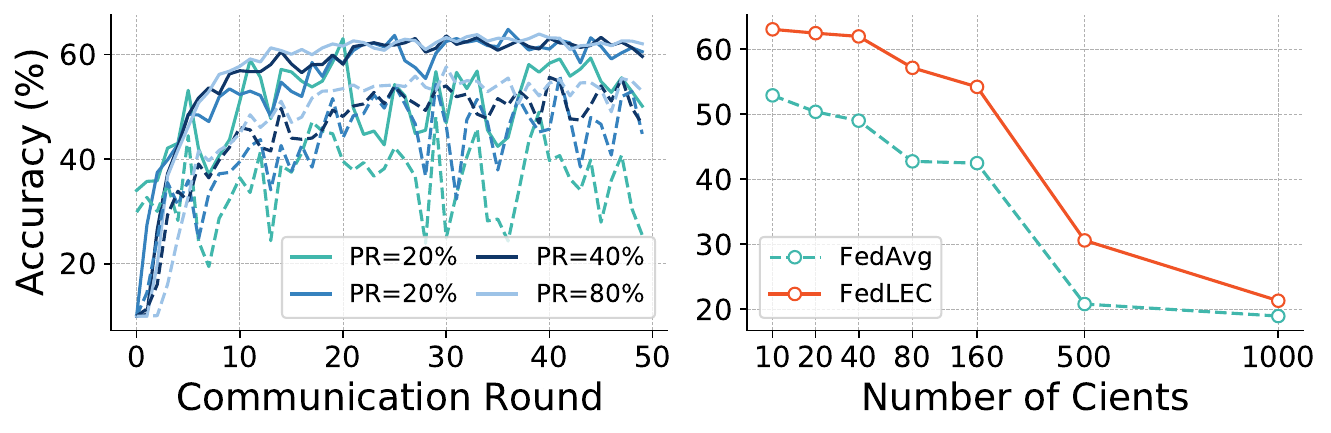} 
\vspace{-0.2in}
\caption{Accuracy patterns of SNN-based FedAvg (dashed line) and FedLEC (solid line) when the participation rate (PR) and the total number of local clients vary under the label skew $p\sim Dir(0.1)$.}
\vspace{-0.1in}
\label{fig:scales}
\end{figure}

\paragraph{Client Participation Rate.} 
In reality, Not all the clients will participate in the entire training process. As the left part of Figure~\ref{fig:scales} illustrates, we simulate this scenario by setting the sample fraction from 20\% to 80\%. 
The FedAvg training curves are less stable than those of FedLEC due to the highly skewed data. The local gradient distributions vary among communication rounds, distorting the correct global model updating direction per round. 

\paragraph{Total Client Number.} 
We study the effect of the number of clients on FedLEC as shown in the right part of Figure~\ref{fig:scales}.
We observe that the overall accuracy trends of both FL algorithms tend to decline as the number of clients increases. 
FedLEC consistently outperforms FedAvg across all large-scale settings, achieving an average improvement of 12.27\%. Although increasing the number of clients reduces the volume of local data per client, which amplifies the risk of biased training, FedLEC effectively mitigates this issue.

\begin{figure}[!t]
\centering
\includegraphics[width=\columnwidth]{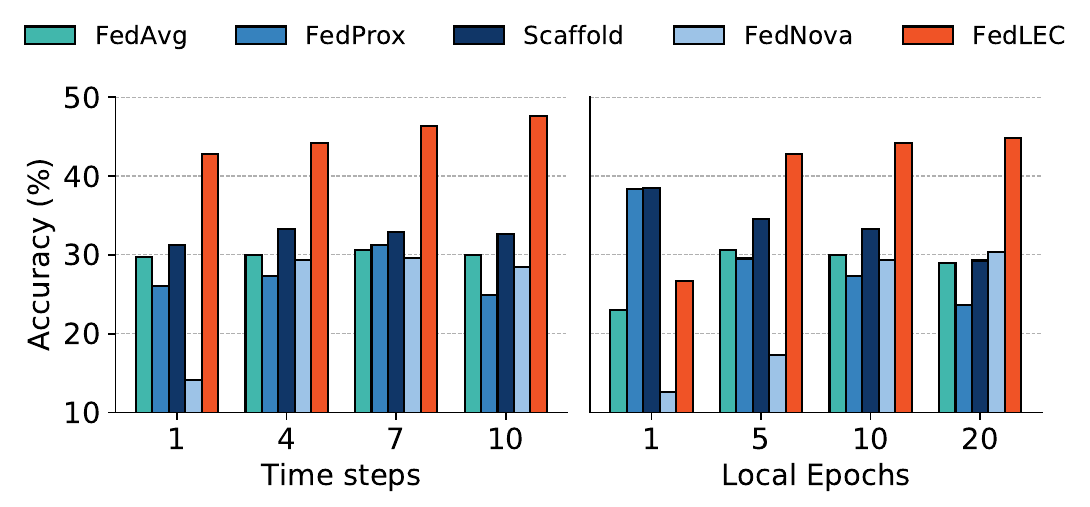} 
\vspace{-0.2in}
\caption{Accuracy patterns of different federated SNN learning algorithms when the time steps and local training epochs vary under the label skew $\#cnum=2$.}
\vspace{-0.1in}
\label{fig:senses}
\end{figure}

\paragraph{Time Steps.} 
The number of time steps dictates the complexity of federated SNN learning. 
We systematically vary this parameter to assess its impact on FedLEC's performance.
Observations in the left part of Figure~\ref{fig:senses} illustrate a continual improvement in performance for FedLEC as time steps increase from 42.74\% to 47.55\%, and the improvement is marginal when the time step increases to 7, as the membrane potential of LIF neurons requires several time steps to stabilize \cite{anumasa2024enhancing} after initialization, allowing the SNN model to generate stable spike trains effectively.

\paragraph{Local Epochs.} 
We vary the local epochs from \{1, 5, 10, 20\} and report the final accuracy under specific label skew in the right part of Figure~\ref{fig:senses}. 
On the one hand, all FL algorithms are sensitive to the number of local epochs. 
When the number of local epochs increases from 1 to 5, FedLEC achieves a more substantial accuracy improvement of 16.10\% compared to other algorithms, as incorporating regularization requires additional epochs for convergence.
On the other hand, the optimal local epoch number varies across FL algorithms under a specific label skew. When local epochs reach 10, the accuracy improvement rate turns marginal for FedLEC and FedNova, whereas algorithms like FedProx and Scaffold might prefer fewer local epochs. 

\paragraph{Privacy Protection Intensity.}

\begin{table}[!t]
\centering
\resizebox{0.9\columnwidth}{!}{%
\begin{tabular}{lrrrrr}
\toprule
\textbf{Intensity $\delta$} & 0.7 & 0.5 & 0.3 & 0.1 & 0 \\ \midrule
\textbf{Accuracy (\%)} & 19.61 & 19.49 & 31.06  & 61.41 & 63.70  \\ \bottomrule
\end{tabular}%
}
\caption{Results of applying local differential privacy technique into FedLEC with various noise intensity $\delta$.}
\label{tab:privacy}
\end{table}

We evaluate the performance of our privacy protection enhanced FedLEC with the local differential privacy strategy. In Particular, we set the noise intensity $\delta=[0, 0.1,0.3, 0.5, 0.7]$ and the corresponding experimental results are shown in Table~\ref{tab:privacy}. We can see that the performance declines as the noise intensity $\delta$ grows, while the performance drop is slight if $\delta$ is not too large. 
When $\delta$ is larger than 0.5, the final accuracy no longer continues to decrease.
Hence, a moderate strength of $\delta$, such as 0.1, is desirable to balance FedLEC accuracy and privacy protection.

\section{Conclusion}

This study addresses the critical challenge of accuracy deterioration in federated SNN learning caused by highly label-skewed data. 
The proposed FedLEC addresses the issue of imbalanced local data distribution by employing a calibrated objective function and mitigates the learning bias exacerbated by missing labels through knowledge distillation for inter-client label alignment.
Extensive experiments on diverse datasets demonstrate that FedLEC significantly outperforms various state-of-the-art FL approaches in most label-skewed scenarios. These results highlight the potential of FedLEC for real-world applications, particularly in energy-constrained and data-heterogeneous edge environments.

\section*{Acknowledgments}
The work of this paper is supported by the National Key Research and Development Program of China under Grant 2022YFB4500100, the National Natural Science Foundation of China under Grant 62125206, the
Zhejiang Provincial Natural Science Foundation of China under 
Grant No. LD24F020014, the National Key Research and Development Program of China No. 2024YDLN0005, and the Regional Innovation and Development Joint Fund of the National Natural Science Foundation of China No. U22A6001.

\bibliographystyle{named}
\bibliography{ijcai25}

\end{document}